\documentclass[letterpaper, 10 pt, conference]{ieeeconf}  

\IEEEoverridecommandlockouts                              

\usepackage{graphics} 
\usepackage{epsfig} 
\usepackage{mathptmx} 
\usepackage{times} 
\usepackage{amsmath} 
\usepackage{amssymb}  
\usepackage{graphicx}
\usepackage{soul}
\usepackage{xcolor}
\sethlcolor{yellow}
\usepackage{dsfont}
\usepackage{xcolor} 
\usepackage{booktabs}
\usepackage{makecell}
\usepackage{multirow}
\usepackage{hyperref}

\title{\LARGE \bf
Variable-Friction In-Hand Manipulation for Arbitrary Objects via Diffusion-Based Imitation Learning
}

\author{Qiyang Yan$^{1}$, Zihan Ding$^{2}$, Xin Zhou$^{1}$, and Adam J. Spiers$^{1}$
\thanks{*Research supported by Imperial College London internal funds}
\thanks{$^{1}$Manipulation and Touch Lab, Department of Electrical and Electronic Engineering, Imperial College London, UK {\tt\small a.spiers@imperial.ac.uk}}
\thanks{$^{2}$Electrical and Computer Engineering Department, Princeton University, US.}%
}

\begin{document}

\maketitle
\thispagestyle{empty}
\pagestyle{empty}


\begin{abstract}
Dexterous in-hand manipulation (IHM) for arbitrary objects is challenging due to the rich and subtle contact process. Variable-friction manipulation is an alternative approach to dexterity, previously demonstrating robust and versatile 2D IHM capabilities with only two single-joint fingers. However, the hard-coded manipulation methods for variable friction hands are restricted to regular polygon objects and limited target poses, as well as requiring the policy to be tailored for each object. This paper proposes an end-to-end learning-based manipulation method to achieve \textit{arbitrary object} manipulation for \textit{any target pose} on real hardware, with minimal engineering efforts and data collection. The method features a diffusion policy-based imitation learning method with co-training from simulation and a small amount of real-world data. With the proposed framework, arbitrary objects including polygons and non-polygons can be precisely manipulated to reach arbitrary goal poses within 2 hours of training on an A100 GPU and only 1 hour of real-world data collection. The precision is higher than previous customized object-specific policies, achieving an average success rate of 71.3\% with average pose error being 2.676 mm and 1.902$^\circ$. Code and videos can be found at: \href{https://sites.google.com/view/vf-ihm-il/home}{https://sites.google.com/view/vf-ihm-il/home}.

\end{abstract}

\section{INTRODUCTION}

Humans effortlessly perform in-hand manipulation (IHM) for everyday tasks, such as reorienting and repositioning arbitrary complex objects like pens or keys without re-grasping \cite{manipulation_behavior}. These actions involve controlled transitions and re-orientations that are crucial but challenging for robots. While impressive IHM tasks have been performed by dexterous hands through learning-based methods \cite{openai_hand, mit_hand, il_human_hand, il_tilde}, those platforms tend to suffer from mechanical complexity. This complexity translates to extremely high hardware and maintenance costs as well as control complexity. 

\begin{figure}[t]
    \centering
    \includegraphics[width=0.9\linewidth]{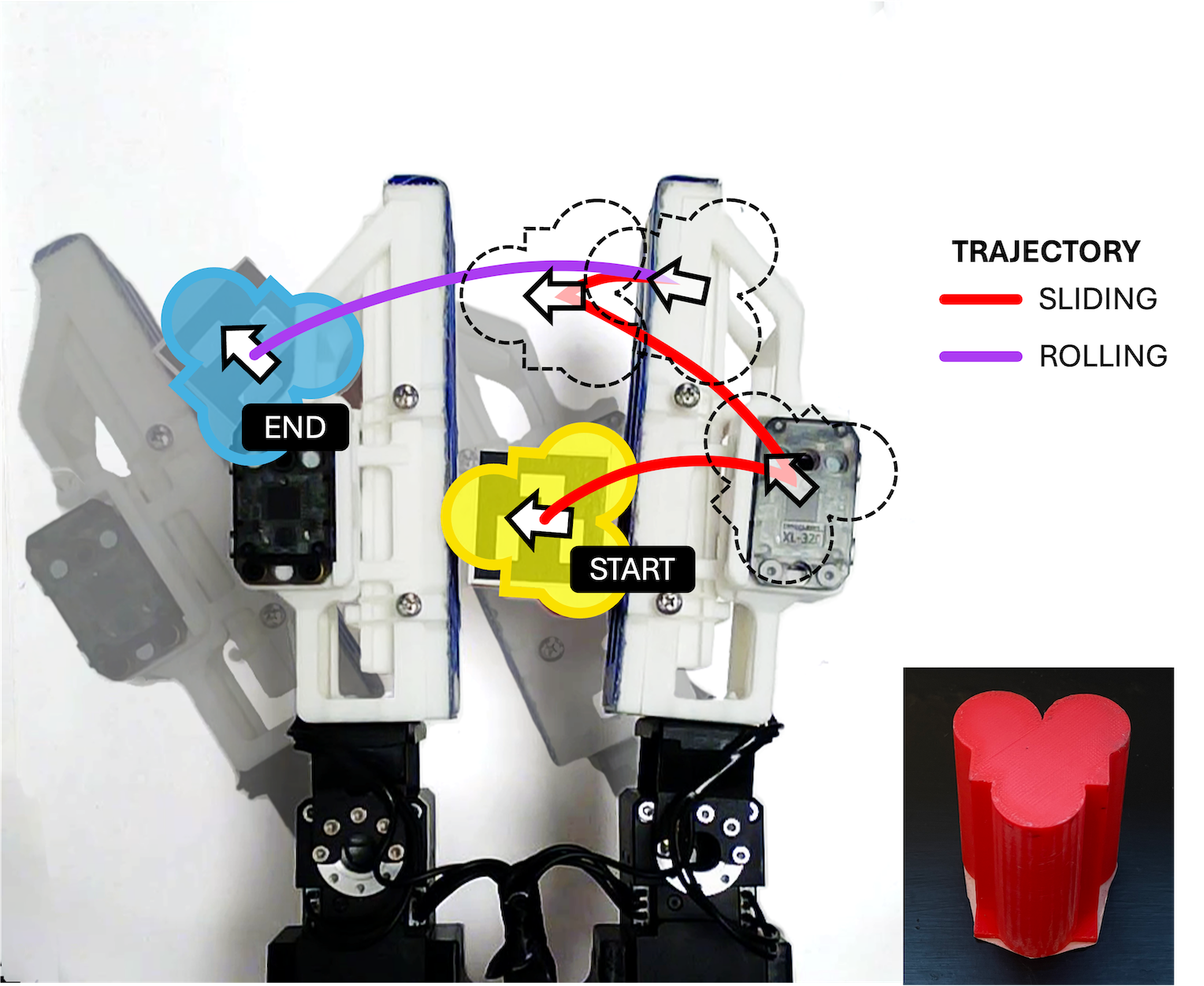}
    \caption{The manipulation trajectory of a complex Three-Cylinder object using the learned control policy. An Aruco marker obscures the body of the object, which is why the shape has been superimposed.
    }
    \label{fig:manipulation_demo}
    \vspace{-.3in}
\end{figure}

In comparison, the 2-DOF variable-friction (VF) hand of \cite{vff} can achieve high object-dexterity despite its simple morphology. Inspired by the ability of human fingertips to selectively slide and grip objects, the VF-hand can achieve the same by actively switching its finger surface between high and low friction states. This allows objects to be slid or rolled along the finger surfaces.

\begin{figure*}[t]
    \centering
    \includegraphics[width=0.95\linewidth]{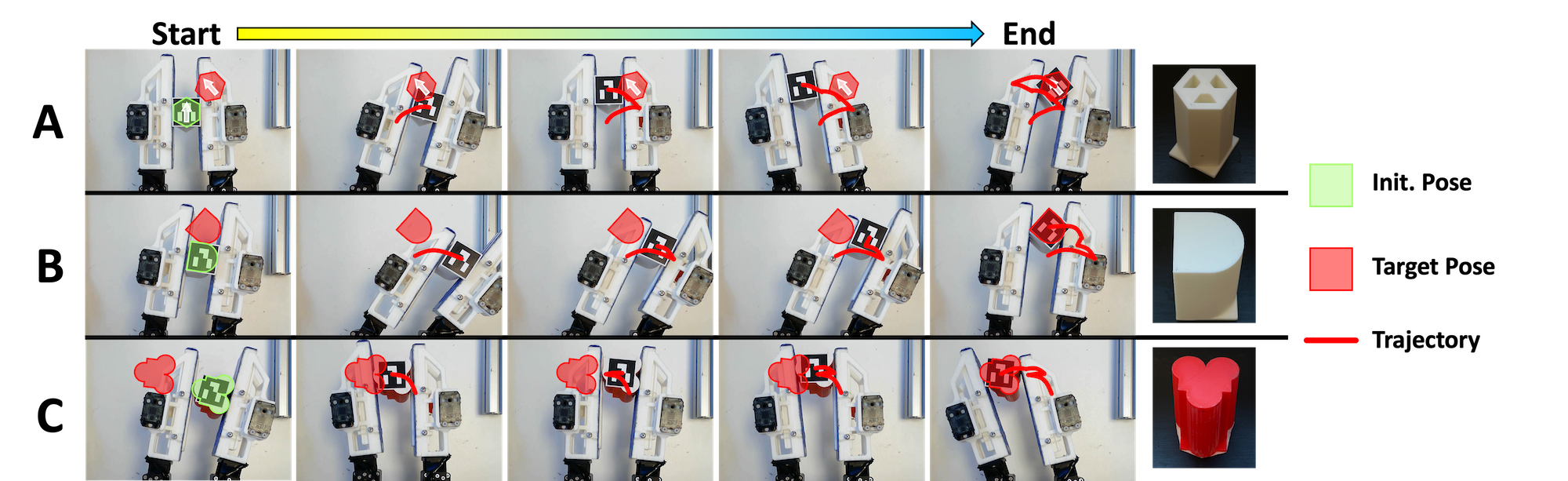}
    \caption{In-hand manipulation of complex objects using the variable-friction hand, trained through our diffusion-based co-train imitation learning pipeline.}
    \label{fig:manipulation_process}
    \vspace{-.2in}
\end{figure*}

Several manipulation-planning strategies have been developed for VF grippers \cite{vff_control, vff_control_region}. The current state-of-the-art is a hybrid of planning and visual servoing \cite{vff_control} via an offline A* pathfinding algorithm to determine the most efficient trajectory through the combination of hard-coded model-based manipulation `primitives'. While this technique demonstrates the potential of variable friction hands, it also suffers from several limitations: (a) The model-based controller restricts manipulation to simple polygonal objects, such as cube and hexagonal prisms. (b) Rotational primitives only allow rolling an object from one flat surface to another. (c) Target poses (orientation and location) are restricted to positions where a flat surface of the object is in contact with a finger. 

We address these shortcomings via a new learning-based approach for VF-manipulation. An end-to-end data-efficient learning framework has been developed, based on imitation learning with diffusion policy. This enables the VF gripper to manipulate complex object shapes from arbitrary initial poses to arbitrary target poses in a continuous manipulation range. While our work focuses on the VF gripper, we believe this methodology is well suited to other uncommon gripper morphologies with non-holonomic object motion constraints.

In addition, we determined that imitation learning with diffusion policy \cite{diffusion_policy} can effectively learn high-precision policies, even with demonstrations generated via an imprecise RL policy optimized for manipulation smoothness. Finally, we have demonstrated that co-training the diffusion policy with a mix of real and simulated data increases both task success rate and resource efficiency. 


\section{Related Work}

Reinforcement learning (RL) learns optimal policies through interactions, and has shown promising results for in-hand manipulation (IHM) in recent years. Some work utilizing model-based approaches with explicit environment representations \cite{learned_local_models, dynamic_models,model_rl}, while model-free approaches learn policies directly \cite{general_system_IHM,openai_hand,openAI_2}. IHM RL policy training is commonly performed in simulation \cite{ding2023learning}, as collecting enough interaction samples in the real world is often too resource intensive. However, training in simulation still requires a large amount of computing power (usually performed on CPU / GPU clusters) and training time due to IHM's task complexity. Imitation Learning (IL) is a frequently-used method to reduce the required training effort for learned IHM policies \cite{il_survey} with growing popularity for IHM applications \cite{IHM_survey}. In addition, the behavior cloning in IL does not require carefully engineered simulation environments (no reward function is needed and it is more forgiving regarding action \& observation space). Robot hands trained via this method learn from expert demonstrations, often generated via observing human object interactions \cite{deep_rl,deep_rl2,il_human_hand,arunachalam2023dexterous}, teleoperation systems \cite{ACT,bunny}, RL policies \cite{mit_hand}, or hand-crafted control strategies \cite{il_hand_craft}.

Recently, IL with the diffusion policy \cite{diffusion_policy,ding2023consistency} (which generates manipulation behaviour from demonstrations via a denoising diffusion process) has shown to be very effective. General tasks such as object-transportation or reorientation \cite{diffusion_policy}, and even complex domestic tasks such as cooking or chair rearranging \cite{mobileAloha} have been successfully tested with this approach. In the context of IHM, \cite{il_tilde} presented a complex hand with 4 fingers, each consisting of a delta mechanism, which used the diffusion policy to perform basic tasks such as cap twisting and syringe pushing.

\section{Problem Formulation}

\subsection{Imitation Learning with Diffusion Policy}
 The manipulation policy is formulated as a Denoising Diffusion Probabilistic Model (DDPM) \cite{ho2020denoising}, which effectively captures the multi-modal distribution through a denoising process. For matching data distribution with probability $p(x_0)$, the forward diffusion process iteratively adds Gaussian noise $\epsilon$ to the data sample such that at step $k$, we have $x_k = \sqrt{\bar{\alpha}_k} x_0 + \sqrt{1 - \bar{\alpha}_k} \epsilon, \quad 1 \leq k \leq K$
with noise $\epsilon \sim \mathcal{N}(0, \mathbf{I})$ and $\bar{\alpha}_k=\Pi_{i=1}^k \alpha_i$ being a predetermined noise schedule. The reverse denoising process follows:
\begin{align}
     x_{k-1}=\frac{1}{\sqrt{\alpha_k}}x_k-\frac{1-\alpha_k}{\sqrt{1-\bar{\alpha}_k}\sqrt{\alpha_k}}\epsilon_\theta+\mathcal{N}(0, \sigma^2_k I)
    \label{eq:x_k_1}
\end{align}
as the posterior distribution of Gaussian following the Bayes' rule, with variance $\sigma_k=\frac{(1-\alpha_k)(1-\bar{\alpha}_{k-1})}{1-\bar{\alpha}_k}$. $\epsilon_\theta$ is the predicted noise by a model parameterized by $\theta$, which can be optimized with the following diffusion loss:
{
\small
\begin{align}
    \mathcal{L}=\mathbb{E}_{x_0\sim\mathcal{D}, \epsilon\sim\mathcal{N}(0, \mathbf{I}),k\in\{1,\dots,K\}}(\epsilon - \epsilon_\theta(\sqrt{\bar{\alpha}_k} x_0 + \sqrt{1 - \bar{\alpha}_k} \epsilon, k))^2
    \label{eq:diffusion_loss}
\end{align}
}

The diffusion policy use DDPM to approximate the conditional distribution $p(a|o)$, \emph{i.e.}, predicting action $a$ conditioned on observation $o$.
For imitation learning, the dataset $\mathcal{D}=\{(o,a)\}$ is pre-collected with a given behavior policy $\pi^b$, and the current diffusion policy $\pi_\theta(a|o)$ is optimized to minimize the distribution divergence from the behavior policy by Eq.~\eqref{eq:diffusion_loss}.

\subsection{Reinforcement Learning}
For RL, we define a Markov decision process $(\mathcal{O}, \mathcal{A}, R, \mathcal{T}, \gamma)$, where $\mathcal{O}$ is the observation space, $\mathcal{A}$ is the action space, $R(o,a):\mathcal{O}\times \mathcal{A}\rightarrow \mathbb{R}$ is the reward function, $\mathcal{T}(o'|o,a):\mathcal{O}\times \mathcal{A}\rightarrow \Pr(\mathcal{O})$ is the stochastic transition function, and $\gamma\in[0,1]$ is the discount factor for value estimation.  The policy in RL is optimized by maximizing its discounted cumulative reward: $\mathbb{E}_\pi[\sum_{t=0}^\infty \gamma^t r(o_t, a_t)]$.

\section{System Overview}
\subsection{Hardware Setup}
\label{sec:hardware}

\subsubsection{Variable Friction Hand}
\label{sec:hand}
\begin{figure}[t]
    \vspace{0.13cm} 
    \centering
    \includegraphics[width=1\linewidth]{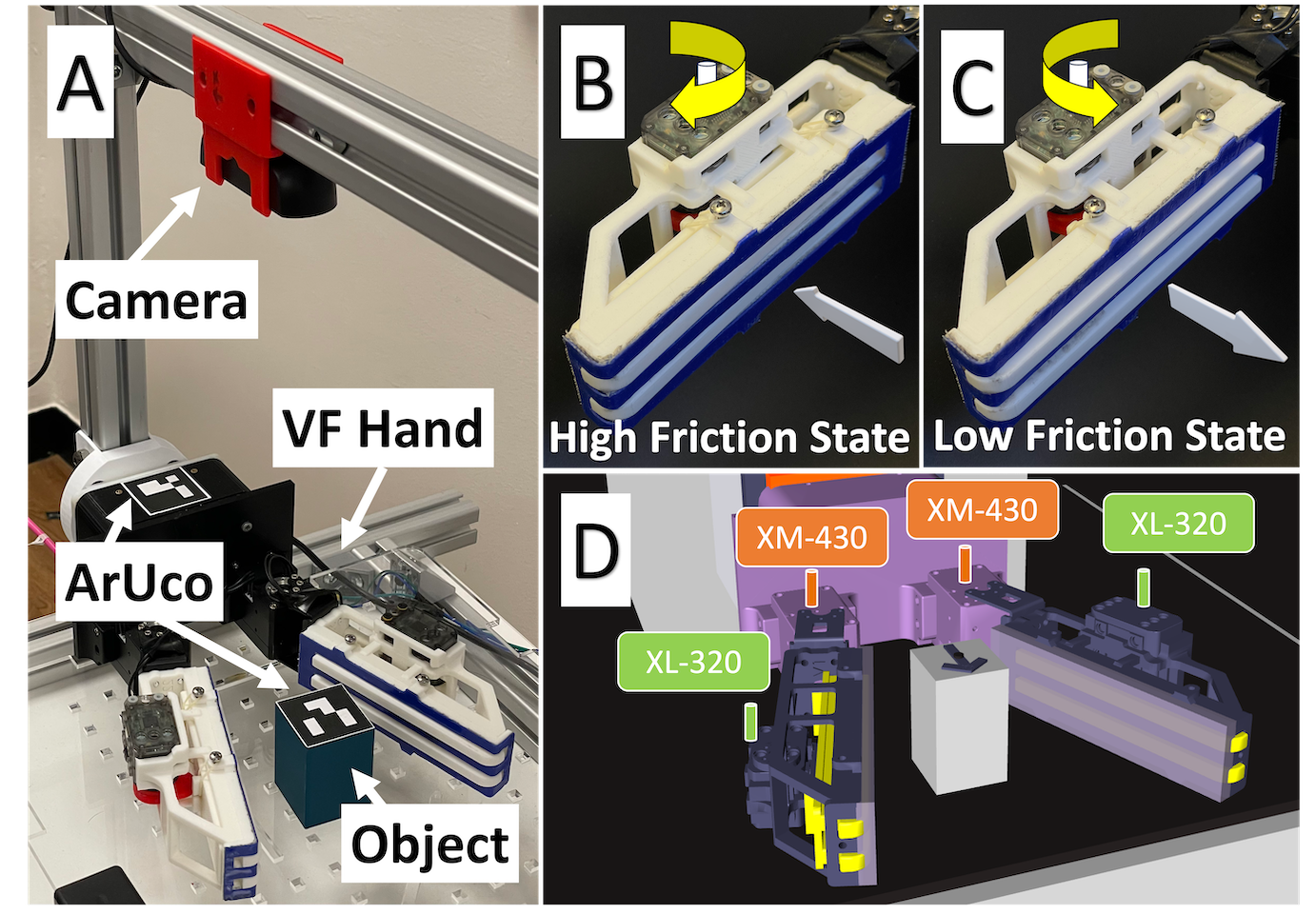}
    \caption{A: The Variable Friction hand mounted on the training \& testing rig. B: When both fingers are in high-friction mode, objects pivot. C: When one of the fingers is in low-friction mode, objects slide along the finger. D. Rendering of the hand and object in the MuJoCo Simulation Environment.}
    \label{fig:hardware}
    \vspace{-.2in}
\end{figure}

The VF hand in this work builds on the design of \cite{vff} and is shown in Fig.~\ref{fig:hardware}. This hand is 3D printed in PLA material. Two Dynamixel XM-430 servo motors actuate the two rotary fingers, while two lower-cost Dynamixel XL-320 servos switch between low- and high-friction finger surfaces via a pulley-cam mechanism described in \cite{vff}. Objects are manipulated via a `push-pull' control scheme: the pulling finger is in position control mode and `pulls away' from the object, whilst the pushing finger is in current control mode, pushing the object firmly against the other finger with a constant torque. During this motion, objects slide along the finger surface if the finger is in low-friction mode (Fig.~\ref{fig:hardware}B); if in high-friction mode, objects typically pivot around the contact point (Fig.~\ref{fig:hardware}C). This behaviour is demonstrated in the accompanying video.

The variable friction hand is mounted on an aluminum strut frame shown in Fig.~\ref{fig:hardware}A. An optional transparent acrylic board was utilized during experiments such that all ArUco markers at the same height, ensuring visual tracking reliability. Our experiments verify the policy performances with and without object support, as demonstrated on the website. A generic 1080p webcam is mounted on the aluminium strut above the hand and is used for tracking object poses. An AruCo marker \cite{aruco} is placed on each manipulated object, and its pose (X and Y location \& Z orientation) is inferred relative to a reference AruCo marker placed on the hand's base. The visual tracking system is calibrated and implemented following \cite{aruco}. The tracking system was evaluated to have a mean absolute error of approximately 1 mm with a standard deviation of 0.75 mm.

\subsection{Simulation Environment}
\label{sec:sim_env}

The simulation environment, built in the MuJoCo engine, features the 2-DoF Variable Friction Hand and objects for manipulation and visualization (see Fig. \ref{fig:hardware}D). 

To simulate the friction pad’s properties, MuJoCo’s soft contact model is hardened by adjusting solver parameters (solref, solimp), using an elliptical friction cone, Newton solver, and a high impratio. The Noslip solver (optional) ensures no slip, matching the real gripper’s behavior. The MultiCCD flag is enabled for more stable contact with meshes at an additional computational cost.


To minimize the sim-to-real gap, we set the physical, geometric, and dynamics parameters through system identification (SI)~\cite{abbeel2005exploration} by comparing the simulation and real trajectories as in \cite{valassakis2020crossing}. Domain randomization~\cite{dr, valassakis2020crossing} has been applied to parameters in the policy training and data collection process, with their ranges identified during SI.


\section{Sim-Real Co-training Framework}
\label{sec:method_high_level}
\begin{figure*}[t]
\vspace{0.05cm} 

    \centering
    \includegraphics[width=0.8\linewidth]{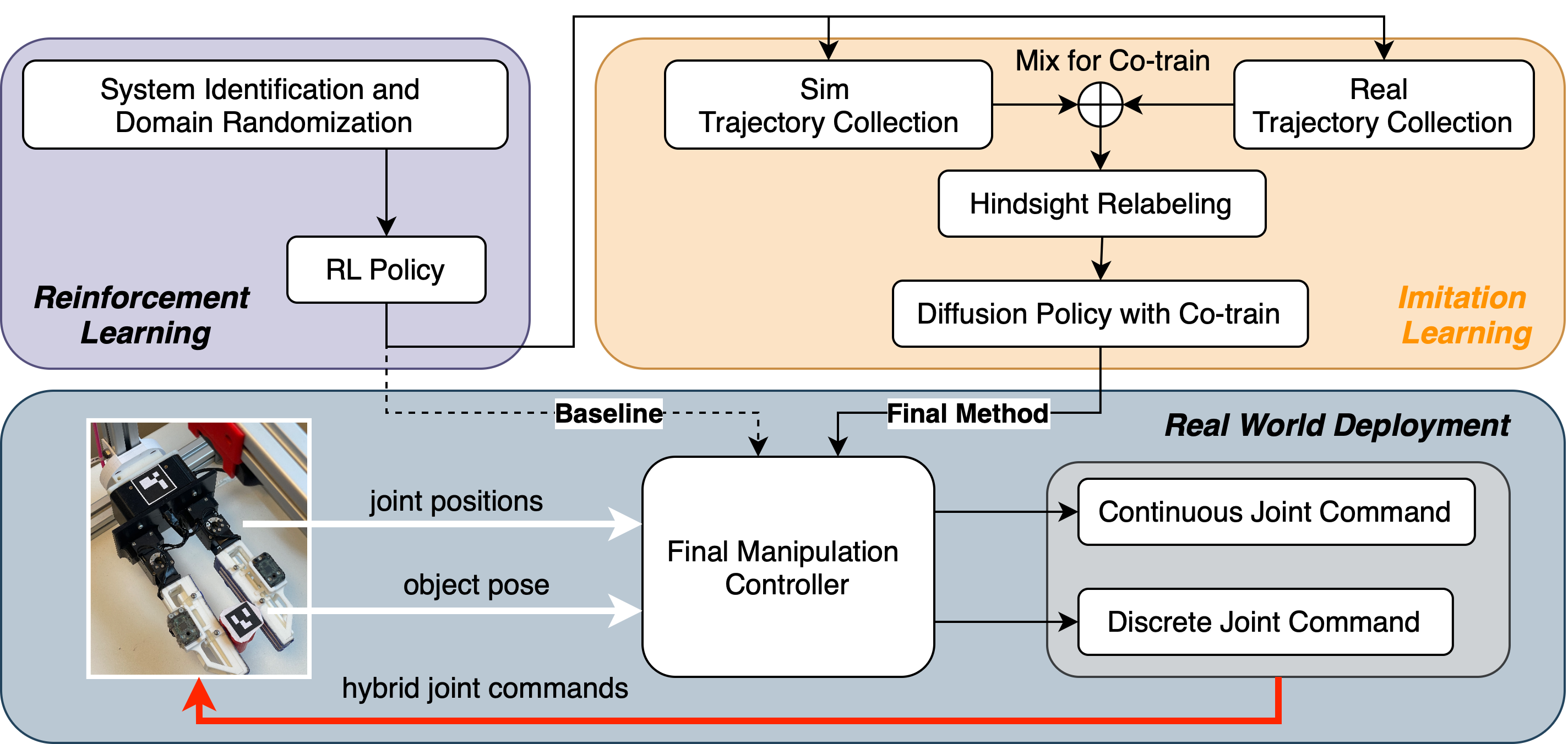}
    \caption{Training Framework of our Co-Train IL method. The IL policy is represented as a diffusion model, and trained with a mix of simulation and real data. The RL policy is used to generate demonstrations for IL, but also used as a baseline during performance analysis.}
    \label{Training Framework}
    \vspace{-.1in}
\end{figure*}

Our initial experiments testified the effectiveness of RL with careful reward engineering and domain randomization for a single regular polygon (a cube). However, as displayed in Fig~\ref{fig:manipulation_demo}, the successful trajectory for variable friction in-hand manipulation does not follow a minimal straight-line distance principle. Hence, the reward function had to be designed specifically for each regular polygon object according to the individual contact surface and interior angle for precise IHM. Such reward shaping is infeasible for non-polygons.

For efficiently manipulating arbitrary objects towards any specified target pose, we propose the imitation learning approach outlined in Fig. \ref{Training Framework}. To capture the multi-modality in the behaviour policy dataset, we adopt the diffusion model for policy representation following~\cite{diffusion_policy}. The benefits of our IL pipeline over the baseline RL approach are manifold: (1) Our straightforward co-training method combines real and simulated demonstrations and effectively bridges the sim-to-real gap. (2) The heavy reward engineering needed for each object under the baseline RL-only approach is avoided. Through \textit{hindsight goal relabeling}~\cite{her}, any smooth trajectory in either simulation or the real world can be relabeled with the actual final pose to compose the demonstration dataset. (3) The restrictions on the object shape and target pose are removed. (4) The IL policy training is significantly more computationally efficient than RL-only policy training (2 hours vs. 15 hours in our experiments).

\subsection{Manipulation Policy Formulation}
The IL and RL manipulation policies $\pi(a|o):\mathcal{O}\rightarrow \Pr(\mathcal{A})$ share a common formulation, which is detailed as follows. 

\paragraph{Observation space} The observation space for RL $o_t \in\mathcal{O}\subset\mathbb{R}^{25}$ consists of the joint state, object state and goal information. For IL, the velocity information in the observation space is zero-padded due to the discrepancy of simulation and real-world velocity.

\paragraph{Hybrid action space} 
The hybrid action space $\mathcal{A} = \mathcal{A}_\text{continuous} \times \mathcal{A}_\text{discrete}$ is designed to allow the VF gripper to perform robust manipulation by switching among discrete friction modes, reducing exploration by focusing on task execution rather than finger coordination. This allows manipulation without support beneath the object even in the presence of sliding.


The discrete action space $\mathcal{A}_\text{discrete}$ specifies the high-level operating mode, specifically in which direction the fingers rotate and which friction modes are used (outlined in Sec.~\ref{sec:hand})). This results in 6 discrete actions (see Table \ref{tab:my_label}). The friction mode switching and finger movements are executed alternatively in each action step to avoid object slipping and motion instability.

\begin{table}[]
\vspace{0.1cm} 

    \centering
    \begin{tabular}{c|c|c|c}
    \hline
        $a_\text{discrete}$ & Action Type & L & R \\ \hline
        0 & Slide up on right finger & PC+HF & TC+LF\\
        1 & Slide down on right finger & TC+HF & PC+LF \\
        2 & Slide up on left finger & TC+LF & PC+HF \\
        3 & Slide down on left finger & PC+LF & TC+HF \\
        4 & Rotate clockwise & TC+HF & PC+HF\\
        5 & Rotate anticlockwise & PC+HF & TC+HF \\ \hline
    \end{tabular}
    \caption{\textbf{Discrete Action Space} L: Left finger. R: Right finger. PC: Position-controlled. TC: Torque-controlled. HF: High-friction. LF: Low-friction}
    \label{tab:my_label}
\vspace{-.3in}
\end{table}

The continuous action space $\mathcal{A}_\text{continuous}$ specifies the relative action $a_t$ between the active finger's current joint angle ($q^{\text{active}}_t$) and its target joint angle ($q^{\text{target}}_t$):

\[ q_{t}^{\text{target}} = q_t^{\text{active}} + a_t \]
The action range is constrained to within $[0, 18.9]$ degrees to limit the finger movement velocity, as a higher $a_t$ at each time step would cause the position-controlled leading finger to accelerate excessively. Note that $a_t$ is never negative, as the leading finger is per definition moving in one direction (the leading finger becomes the following finger once the movement direction is reversed). The control frequency for the above system is 2.5 Hz.



\subsection{Smooth Manipulation via Reinforcement Learning}

\label{sec:hybrid_rl}
 The hybrid action space is handled by a multi-head model architecture for both continuous and discrete prediction. As a key component in RL policy training, the reward function for a precise RL policy required heavy manual design for each object, which was feasible for only regular polygonal objects. Thus we opted for a reward function to obtain a RL policy capable of smoothly manipulating unseen objects with loosened precision requirements, significantly reducing training time.
 


\subsubsection{Reward function}
\label{sec:reward_function}
The task is successful if the Euclidean distance $\Bar{d}$ and angular $\Bar{\theta}$ difference between the object's current and the goal pose is both smaller than the threshold value $\Bar{d}, \Bar{\theta}$, i.e. $\Delta d \leq \bar{d}, \Delta \theta \leq \bar{\theta} $. This comprises the sparse success reward. However, only with sparse reward the policy learning can be less efficient due to lack of dense reward signals.
Considering the gripper's special side-to-side movement pattern (Fig. \ref{fig:manipulation_demo}), we choose to also use the angular coordinates $(r, \theta)$ apart from the Euclidean coordinates for the object pose.  
Specifically, we use
$\Delta r_t = (|r_1-r_{g1}|+|r_2-r_{g2}|)/2$ as the dense reward function, where $r_1$ and $r_2$ are the distances from the two finger bases to the object's current position, while $r_{g1}$ and $r_{g2}$ are the same distances but for the goal position.
We also penalized the agent for manipulating the object out of the legal operation range with a sparse penalty term. The complete reward function is mathematically expressed as:
\begin{align}
r &= c_1 \mathds{1}(|\Delta \theta_t| < \bar{\theta} \ \land \ |\Delta d_t| < \bar{d}) \tag*{\text{sparse success reward}} \\
&- c_2 |\Delta r_t| \tag*{\text{dense task reward}} \\
&- c_3 \mathds{1}(\text{Object out of legal range}) \tag*{\text{sparse penalty}}
\end{align}
where $c1, c2, c3 > 0$ are weighting coefficients, and $\mathds{1}$ is an indicator function that identifies whether the condition within the bracket is satisfied.


\subsubsection{Training Details}
The RL algorithm applied in our experiments is TD3~\cite{td3} with Hindsight Experience Replay~\cite{her}, with parallel environments (44 CPU cores) for data sampling. The same RL training setting is applied to both our IL approach and the baseline RL policy for comparable results.  The exploration policy is trained on the Cube object only, taking roughly 3.5 hours; while the baseline RL is trained for at least 15 hours to converge to a high success rate with the same level of precision for each object. The episodes are terminated if the object does not reach success within performing 10 action steps. This led to efficient and smooth (but imprecise) manipulation policies for arbitrary unseen objects. This smoothness-optimized policy is then used to generate expert demonstrations described further in the next subsection.

\subsection{Collecting Demonstrations with Hindsight Relabeling}
\label{sec:demo}
The demonstration dataset is collected with the smoothness-optimized RL policy in both simulation and reality. We adopt \textit{hindsight goal relabeling}~\cite{her, her_il} to collect demonstration trajectories. The final state of a sampled trajectory is relabeled as the goal state and treated as an expert demonstration. The major benefit of this method is that it does not require the RL policy to be optimal for arbitrary objects and specified targets, as long as the trajectory is smooth within the task domain. The behavior policy is not restricted to a RL policy. 

\subsection{Sim-Real Co-Trained Diffusion Imitation Learning}
\label{sec:il}
For the next stage, we use behaviour cloning with a diffusion policy \cite{diffusion_policy} to generate object-specific precise manipulation policies. We introduce the sim-real co-training framework for training the diffusion policy, using demonstrations generated via the above hindsight goal relabeling approach. For each object, we collect 100 real demos $\mathcal{D}_{real}$ (1 h/object), and 10,000 simulation demos $\mathcal{D}_{sim}$ (30 min/object).

The training objective for the diffusion policy is defined as:
\begin{align}
    \min_{\theta}
&\mathbb{E}_{(o^i, a^i) \sim \mathcal{D}_\text{sim}\bigcup \mathcal{D}_\text{real} }
    \left[
        \mathcal{L}_\theta(o^i, a^i)
    \right]\\
 =\min_{\theta} &\mathbb{E}_{\substack{(o^i, a^i) \sim \mathcal{D}_\text{sim}\bigcup \mathcal{D}_\text{real} \\\epsilon\sim\mathcal{N}(0, \mathbf{I}), k\in\{1,\dots,K\}}}(\epsilon - \epsilon_\theta(\sqrt{\bar{\alpha}_k} a^i + \sqrt{1 - \bar{\alpha}_k} \epsilon, o^i, k))^2
\end{align}
where $o^i$ is the observation consisting of object goal pose, joints and object state, $a^i \in \mathbb{R}^{2}$ aligns with the action space defined for RL, and $\mathcal
L$ is the diffusion loss following Eq.~\eqref{eq:diffusion_loss}. 
$\epsilon_\theta$ is $\theta$-parameterized diffusion policy for noise prediction.
We sample with equal probability from the sim data $\mathcal{D}_\text{sim}$ and real data $\mathcal{D}_\text{real}$ with a batch size of 1024. Observations are normalized based on the statistics of the real data distribution.

\section{Real-World Experiments}
\label{sec:experiment}

\subsection{Experiment setting}
\begin{figure}[t]
    \vspace{0.1cm}
    \centering
    \includegraphics[width=0.8\linewidth]{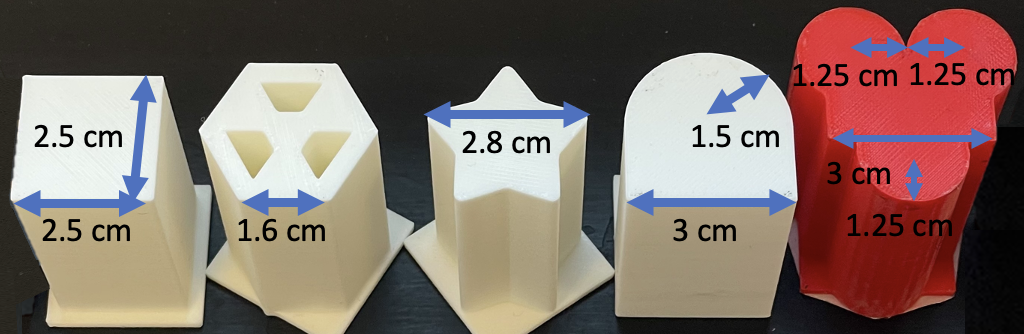}
    \caption{The objects used in the experiments. From left to right: Cube, Hexagon (regular polygons), Star (irregular polygon), Cube Cylinder and Three-Cylinder (non-polygons)}
    \label{fig:Objects}
    \vspace{-.1in}
\end{figure}

As shown in Fig. \ref{fig:Objects}, we designed 5 objects of varying shapes, including regular and irregular polygons, as well as non-polygons. The objects are 3D-printed with the same geometries as their simulation counterparts, but surface properties like friction may differ. The expert demonstrations were collected via the same exploration RL policy trained on the Cube, but final precise IL policies are object-specific. 


Performance was evaluated on the real-world platform outlined in Sec. \ref{sec:hardware}. We average performance across 3 random seeds with 10 trials each and state standard deviations. Three performance measures were used: success rate of objects ending within 5 mm and 5.73 degrees (0.1 radians) from targets with a maximum of 10 action steps; positional error (center-distance to target) for successful episodes; and rotation error for successful episodes.

\subsection{IL Performance Results}
\label{sec:main_results}
\subsubsection{Main Results}
As shown in Table. \ref{tab:experiment1}, our method successfully manipulates the five test objects to arbitrary goal poses sampled within range with a success rate of 71.3\% and an average positional error of 2.676 mm and a rotational error of 1.902 degrees for successful attempts. Notably, the error levels remain consistent across objects of varying complexity, which we attribute to system error. Fig. \ref{fig:co-train evaluation} visualizes these success rates tested on the real-world VF Hand.

\begin{table}[ht]
\centering 
\caption{Final Co-Trained IL Method (10,000 sim and 100 real demos) tested on the VF hand} 
\label{tab:main_result} 
\resizebox{\columnwidth}{!}{
\begin{tabular}{cccc}
\multirow{2}{*}{\textbf{Object}}  & \textbf{Success Rate ($\%$)} & \textbf{Position Error} & \textbf{Rotation Error} \\ 
                &    \textbf{In Reality}                    & \textbf{(mm)}           & \textbf{(degree)}        \\ 
                \cmidrule(lr){1-4}
\textbf{Cube}           & $76.7\pm 20.8$  & $2.550\pm 0.910$ & $1.260\pm 0.998$ \\
\textbf{Hexagon}        & $76.7\pm12.5$  & $2.415\pm1.195$ & $2.105\pm1.259$ \\
\textbf{Star}           & $73.3\pm12.5$   & $2.847\pm1.218$ & $2.153\pm1.301$ \\
\textbf{Cube Cylinder}  & $70.0\pm10.0$  & $2.997\pm1.295$  & $1.432\pm1.216$   \\
\textbf{Three-Cylinder} & $60.0\pm26.5$  & $2.573\pm 1.171$ & $2.561\pm 2.022$\\ \cmidrule(lr){1-4}
\textbf{Average}        & $71.3\pm16.5$  & $2.676\pm1.158$ & $1.902\pm1.359$ \\ \cmidrule(lr){1-4}
\end{tabular}
}
\label{tab:experiment1}
\vspace{-.1in}
\end{table}


We observed that the agent can execute effective trajectories toward unseen target poses sampled in distribution at test time. It achieves the goal pose with at most 4 friction changes in most trials, leading to fast task completion with additional benefits of reduced hardware wear and tear.

\subsubsection{IL vs. Baseline RL Policy}
As an RL-only policy following Sec.~\ref{sec:hybrid_rl} required heavy reward engineering for each object, we only conducted this comparison on the Cube. To achieve high success rates, the RL baseline takes 15 hours of training; while the exploration RL policy for our IL method takes only 3.5 hours. From Table.~\ref{tab:il_rl}, we observe that our final IL method with co-train achieves a higher success rate with similar levels of errors for successful episodes, but requires drastically less training time ($36.7\%$). To manipulate unseen objects, only 2 extra hours of IL are needed as the exploration RL policy is shared.

\begin{table}[h!]
\centering 
\caption{Performance Comparison of reward-engineered RL and our Final IL for Cube Manipulation on the real hand} 
\resizebox{1\columnwidth}{!}{
\begin{tabular}{cccc}
\cmidrule(lr){1-4}
\textbf{Method} & \textbf{Success Rate} & \textbf{Training Duration} &\textbf{Real Error (Pos./Rot.)} \\ \cmidrule(lr){1-4}
\multirow{2}{*}{\textbf{Baseline RL}} &\multirow{2}{*}{66.7\%}  & \multirow{2}{*}{15h} & $(2.094 \pm0.863)$ mm / \\ 
 & & & $(3.032\pm1.814)^\circ$ \\ \cmidrule(lr){1-4} 
\multirow{2}{*}{\textbf{Our Method}}  & \multirow{2}{*}{76.7\%} & \multirow{2}{*}{$3.5\text{h(RL)} + 2\text{h(IL)} $} & $(3.336\pm1.211)$ mm /\\ 
&& &  $(2.163\pm1.645)^\circ$ \\ \cmidrule(lr){1-4}
\end{tabular}
}
\label{tab:il_rl}
\vspace{-.1in}
\end{table}

\subsubsection{IL vs. Model-based Policy}

Previous work for variable friction IHM tasks adopts the A* planner method \cite{vff_control}, where manipulation primitives are hard-coded to achieve only face-to-face rolling for restricted target pose reaching, therefore A* is infeasible for arbitrary target poses. Our IL approach does not have such constraints. In Table~\ref{A* comparison}, we compare the two methods under two settings: for 4 pre-specified goal poses (same as in \cite{vff_control}) and for arbitrary poses. Across the Cube and Hexagon, our method demonstrates a lower positional error compared to A* planner, for even arbitrary goal poses.
\begin{table}[h]
\centering 
\caption{Positional error (mm) of the A* Method and our IL method on the real hand} 
\label{A*_compare} 
\resizebox{\columnwidth}{!}{
\begin{tabular}{ccccc} 
& \multicolumn{2}{c}{\textbf{4 Different Achievable Goals}} & \multicolumn{2}{c}{\textbf{Arbitrary Goal Poses}} \\ \cmidrule(lr){2-3} \cmidrule(lr){4-5}
\textbf{Object} & \textbf{Our Method}  & \textbf{A*} & \textbf{Our Method} & \textbf{A*} \\ \cmidrule(lr){1-5}
\textbf{Cube} & $3.055\pm1.635$ & $3.800\pm0.900$ & $3.336\pm1.211$ & - \\
\textbf{Hexagon} & $2.719\pm1.506$ & $3.800\pm0.500$ & $2.415\pm1.195$ & - \\ 
\cmidrule(lr){1-5}
\end{tabular}
}\label{A* comparison}
\vspace{-.2in}
\end{table}

\subsubsection{Failure Cases}
\label{fail}
On occasions where compounded errors lead to target positions being missed, the gripper is observed to perform an additional sliding action followed by a rotation, often managing to eventually reach the target within the required 10 total action steps. Another type of failure is observed for three-cylinders, where the gripper intended to slide the object toward the gripper base but resulted in object rolling. This type of failure is due to the VF hand's fixed base width: the angles between fingers cannot be independently controlled, leading to non-ideally angled forces that cause rolling behaviour even under low frictions. Such failure can be avoided by introducing an actuated prismatic palm with more control over finger angles such as in \cite{etroll}.




\subsection{The Effects of Co-Training}

\subsubsection{Ablation Study -- Determining Real-Data Efficiency}
\begin{figure}[b]
    \centering
    \includegraphics[width=1\linewidth]{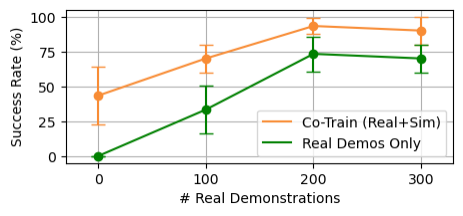}
    \caption{Real robot success rate for using different amounts of real-world demonstrations for object Cube Cylinder.}
    \label{fig:data_scarcity}
    \vspace{-.1in}
\end{figure}

Real data collection usually requires intensive human involvement which is expected to be minimized in the sim-real co-training framework in the pipeline (Sec.~\ref{sec:il}). By keeping the simulation data fixed as 10,000 trajectories and varying the amount of real data demonstrations, we investigate the real data efficiency in the co-training framework. The amount of real data is in $[0, 100, 200, 300]$ trajectories, with a real-world collection speed of 100 trajectories per hour on the real IHM system. The object used for this experiment is the Cube Cylinder. 
The results averaged across three random seeds are shown in Fig.~\ref{fig:data_scarcity}. This indicates that the success rate increases along with the usage of more real data, but plateaus after 200 trajectories, with highest success rate $93.3\%$ for co-training and $73.3\%$ for real data only. 100 trajectories are selected as the default amount in our main experiments in Sec.~\ref{sec:main_results} and \ref{sec:cotrain}, due to a good balance of a low labor-intensive real data collection effort and high success rate of 70\% (a more than 2-fold improvement from 33.3\%). The simulation data also provides significant performance boosts across different real data amounts. This testifies the effectiveness of the proposed co-training framework with a mix of simulation and real data. 
\subsubsection{Co-Training vs. Real/Sim Demos Only}
\label{sec:cotrain}
\begin{figure}[t]
\vspace{0.1cm} 
    \centering
    \includegraphics[width=0.99\linewidth]{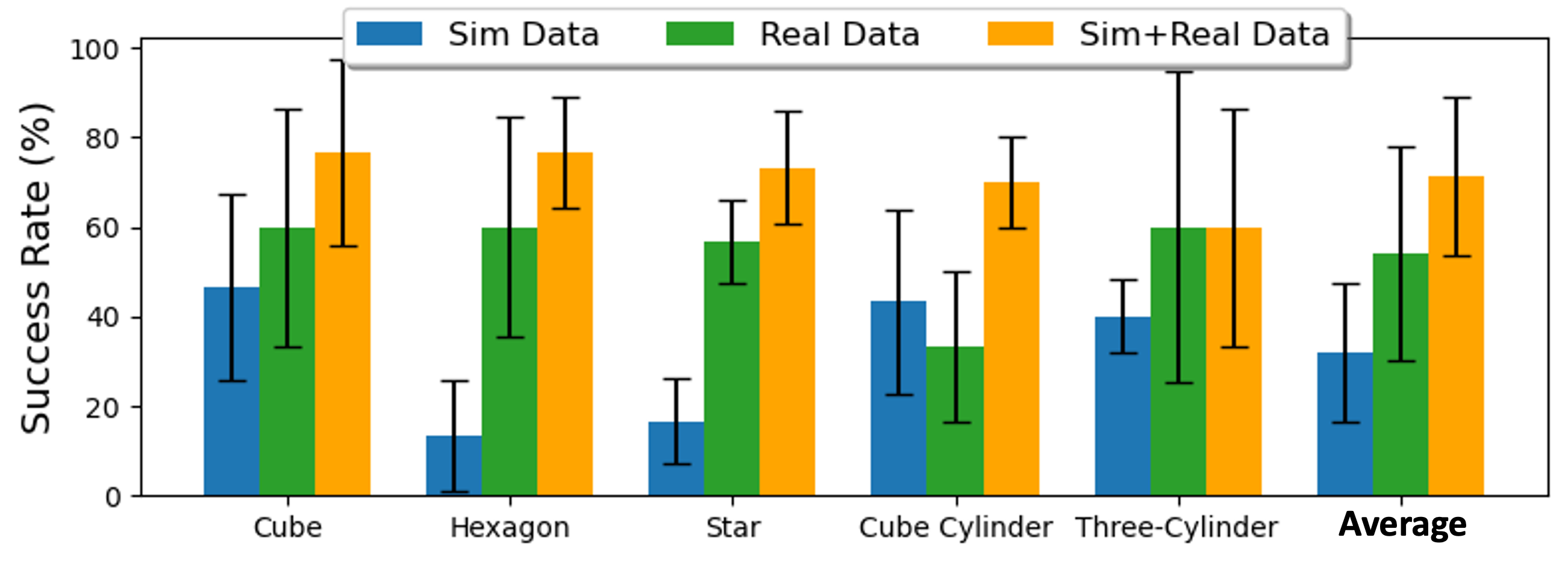}
    \caption{Co-train compared to sim/real demonstrations only, tested on the real-world VF hand.}
    \label{fig:co-train evaluation}
\vspace{-.1in}
\end{figure}
The results in Fig. \ref{fig:co-train evaluation} demonstrate the effectiveness of co-training which leverages both simulation and real-world data. The success rate increases significantly for most objects when co-training is applied, with the Three-Cylinder object being an exception, whose shape was too challenging for the gripper's mechanical design to handle as explained in Sec.\ref{fail}. On average across five objects, the success rate improved from 32.0\% (simulation only) and 54.0\% (real data only) to 71.3\% when using both. Additionally, the standard deviation is reduced for most tasks, indicating greater stability and consistency in performance. 

\subsubsection{Co-Training vs. Fine-tuning}
We also made a comparison with fine-tuning using the cube cylinder by first training the diffusion policy on the simulation demonstrations and then fine-tuning it with the real demonstrations. The results for this experiment use 200 episodes real data. Fine-tuning achieved a success rate of $76.7\%$, with a lower success rate and higher standard deviation compared to co-training. Our conjecture is that fine-tuning can lead to overfitting on few real data easily thus losing its generalization on new poses. These findings confirm that co-training effectively enhances task success while improving reliability across different object manipulations.
\begin{table}[h!]
\centering 
\caption{Comparison of Co-Train and Fine-Tune on the real hand} 
\label{tab:example_table}
\resizebox{\columnwidth}{!}{
\begin{tabular}{cccc}
\cmidrule(lr){1-4}
 & \textbf{Real Data Only} & \textbf{Co-train}  & \textbf{Finetune}\\ \cmidrule(lr){1-4}
\textbf{Success Rate ($\%$)}  & $73.3\pm17.3$ & $93.3\pm5.8$ & $76.7\pm12.5$ \\
\cmidrule(lr){1-4}
\end{tabular}
}
\vspace{-.1in}
\end{table}




\section{Conclusion}

In summary, our method unlocks the VF gripper's IHM capabilities. We trained an automated demonstration generation agent through reinforcement learning, allowing effortless demonstration collection in both simulation and the real world. By deploying diffusion policies co-trained with simulation data, we addressed the scarcity of real-world data and reduced the simulation-to-real-world discrepancy, allowing learning to manipulate arbitrary objects for arbitrary start and goal pose with 2-hours training and 1 hour real-world data collection. Future work includes a general policy for arbitrary objects, through mixing training demonstrations across sample objects. 

\clearpage

\bibliographystyle{IEEEtran}
\bibliography{ref.bib}

\vspace{12pt}

\end{document}